\documentclass[11pt,a4paper]{article}
\usepackage[hyperref]{naaclhlt2018}
\usepackage{times}
\usepackage{latexsym}
\usepackage{amsmath}
\usepackage{bm}
\usepackage{comment}
\usepackage{graphicx}

\usepackage{url}

\aclfinalcopy 
\def\aclpaperid{739} 

\title{Contextual Augmentation:\\Data Augmentation by Words with Paradigmatic Relations}

\author{Sosuke Kobayashi \\
  Preferred Networks, Inc.\\
  {\tt sosk@preferred.jp}}

\date{}

\begin{document}
\maketitle
\begin{abstract}
  We propose a novel data augmentation for labeled sentences called {\it contextual augmentation}.
  We assume an invariance that sentences are natural even if the words in the sentences are replaced with other words with paradigmatic relations.
  We stochastically replace words with other words that are predicted by a bi-directional language model at the word positions.
  Words predicted according to a context are numerous but appropriate for the augmentation of the original words.
  Furthermore, we retrofit a language model with a label-conditional architecture,
  which allows the model to augment sentences without breaking the label-compatibility.
  Through the experiments for six various different text classification tasks,
  we demonstrate that
  the proposed method improves classifiers based on the convolutional or recurrent neural networks.
\end{abstract}

\section{Introduction}

Neural network-based models for NLP have been growing with state-of-the-art results in various tasks,
e.g., dependency parsing~\cite{dyer15}, text classification~\cite{socher13,kim14},
machine translation~\cite{sutskever14}.
However, machine learning models often overfit the training data by losing their generalization.
Generalization performance highly depends on the size and quality of the training data and regularizations.
Preparing a large annotated dataset is very time-consuming.
Instead, automatic data augmentation is popular, particularly in the areas of
vision~\cite{simard98,krizhevsky12,szegedy15}
and speech~\cite{jaitly15,ko15}.
Data augmentation is basically performed based on human knowledge on invariances, rules, or heuristics, e.g., ``even if a picture is flipped, the class of an object should be unchanged''.

\begin{figure}[!t]
  \begin{center}
    \includegraphics[width=7.4cm,clip]{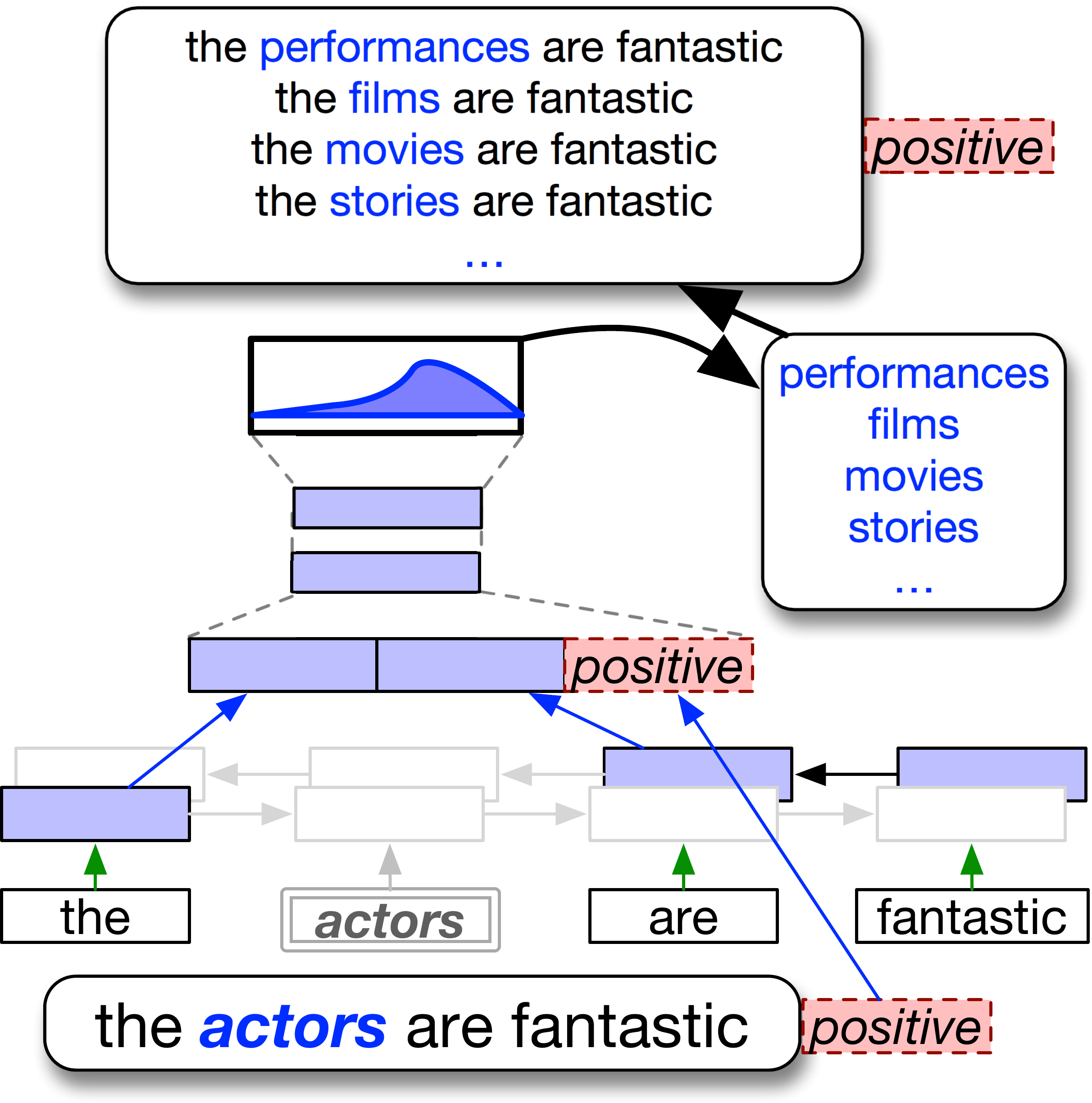}
    \caption{
    Contextual augmentation with a bi-directional RNN language model, when a sentence {\it ``the actors are fantastic''} is augmented by replacing only {\it actors} with words predicted based on the context.
}
    \label{fig:contextual_aug_encode}
  \end{center}
\end{figure}

However, usage of data augmentation for NLP has been limited.
In natural languages, it is very difficult to obtain universal rules for transformations which assure the quality of the produced data and are easy to apply automatically in various domains.
A common approach for such a transformation is to replace words with their synonyms selected from a handcrafted ontology such as WordNet~\cite{miller95,zhang15} or word similarity calculation~\cite{wang15}.
Because words having exactly or nearly the same meanings are very few, synonym-based augmentation can be applied to only a small percentage of the vocabulary.
Other augmentation methods are known but are often developed for specific domains with handcrafted rules or pipelines, with the loss of generality.

In this paper, we propose a novel data augmentation method called {\it contextual augmentation}.
Our method offers a wider range of substitute words by using words predicted by a bi-directional language model (LM) according to the context, as shown in Figure~\ref{fig:contextual_aug_encode}.
This contextual prediction suggests various words that have paradigmatic relations~\cite{saussure1916} with the original words.
Such words can also be good substitutes for augmentation.
Furthermore, to prevent word replacement that is incompatible with the annotated labels of the original sentences, we retrofit the LM with a label-conditional architecture.
Through the experiment, we demonstrate that
the proposed conditional LM produces good words for augmentation,
and contextual augmentation improves classifiers using recurrent or convolutional neural networks (RNN or CNN) in various classification tasks.

\section{Proposed Method}

For performing data augmentation by replacing words in a text with other words, prior works~\cite{zhang15,wang15} used synonyms as substitute words for the original words.
However, synonyms are very limited and the synonym-based augmentation cannot produce numerous different patterns from the original texts.
We propose {\it contextual augmentation}, a novel method to augment words with more varied words.
Instead of the synonyms, we use words that are predicted by a LM given the context surrounding the original words to be augmented,
as shown in Figure~\ref{fig:contextual_aug_encode}.

\subsection{Motivation}

First, we explain the motivation of our proposed method by
referring to an example with a sentence from the Stanford Sentiment Treebank (SST)~\cite{socher13},
which is a dataset of sentiment-labeled movie reviews.
The sentence, {\it ``the actors are fantastic.''}, is annotated with a positive label.
When augmentation is performed for the word (position) {\it ``actors''},
how widely can we augment it?
According to the prior works, we can use words from a synset for the word {\it actor} obtained from WordNet ({\it histrion, player, thespian,} and {\it role\_player}).
The synset contains words that have meanings similar to the word {\it actor} on average.\footnote{
Actually, the word {\it actor} has another synset containing other words such as {\it doer} and {\it worker}.
Thus, this synonym-based approach further requires word sense disambiguation or some rules for selecting ideal synsets.}
However, for data augmentation, the word {\it actors} can be further replaced with non-synonym words such as
{\it characters, movies, stories,} and {\it songs} or various other nouns, while retaining the positive sentiment and naturalness.
Considering the generalization, training with maximum patterns will boost the model performance more.

We propose using numerous words that have the paradigmatic relations with the original words.
A LM has the desirable property to assign high probabilities to such words,
even if the words themselves are not similar to the original word to be replaced.

\subsection{Word Prediction based on Context}

For our proposed method, we requires a LM
for calculating the word probability at a position $i$
based on its context.
The context is a sequence of words surrounding an original word $w_i$ in a sentence $S$, i.e., cloze sentence $S\backslash\{w_i\}$.
The calculated probability is $p(\cdot|S\backslash\{w_i\})$.
Specifically,
we use a bi-directional LSTM-RNN~\cite{Hochreiter97} LM.
For prediction at position $i$,
the model encodes the surrounding words individually rightward and leftward (see Figure~\ref{fig:contextual_aug_encode}).
As well as typical uni-directional RNN LMs, the outputs from adjacent positions are used for calculating the probability at target position $i$.
The outputs from both the directions are concatenated and fed into the following feed-forward neural network, which produces words with a probability distribution over the vocabulary.

In contextual augmentation, new substitutes for word $w_i$ can be smoothly sampled from a given probability distribution,
$p(\cdot|S \backslash \{w_i\})$, while prior works selected top-K words conclusively.
In this study, we sample words for augmentation at each update during the training of a model.
To control the strength of augmentation,
we introduce temperature parameter $\tau$
and use an annealed distribution
$p_{\tau}(\cdot|S \backslash \{w_i\}) \propto p(\cdot|S\backslash\{w_i\})^{1/\tau}$.
If the temperature becomes infinity ($\tau \to \infty$), the words are sampled from a uniform distribution.~\footnote{
Bengio et al.~\shortcite{bengio15} reported that stochastic replacements with uniformly sampled words improved a neural encoder-decoder model for image captioning.
}
If it becomes zero ($\tau \to 0$), the augmentation words are always words predicted with the highest probability.
The sampled words can be obtained at one time at each word position in the sentences.
We replace each word simultaneously with a probability as well as Wang and Yang~\shortcite{wang15} for efficiency.

\subsection{Conditional Constraint}

Finally, we introduce a novel approach to address the issue that
context-aware augmentation is not always compatible with annotated labels.
For understanding the issue,
again, consider the example, {\it ``the actors are fantastic.''}, which is annotated with a positive label.
If contextual augmentation, as described so far, is simply performed for the word (position of) {\it fantastic},
a LM often assigns high probabilities
to words such as {\it bad} or {\it terrible} as well as {\it good} or {\it entertaining}, although they are mutually contradictory to the annotated labels of positive or negative.
Thus, such a simple augmentation can possibly generate sentences that are implausible with respect to their original labels and harmful for model training.

To address this issue,
we introduce a conditional constraint
that controls the replacement of words to prevent the generated words from reversing the information related to the labels of the sentences.
We alter a LM
to a label-conditional LM,
i.e., for position $i$ in sentence $S$ with label $y$, we aim to calculate $p_{\tau}(\cdot | y, S \backslash \{w_i\})$ instead of the default $p_{\tau}(\cdot | S \backslash \{w_i\})$ within the model.
Specifically, we concatenate each embedded label $y$ with a hidden layer of the feed-forward network in the bi-directional LM,
so that the output is calculated from a mixture of information from both the label and context.

\section{Experiment}

\subsection{Settings}

We tested combinations of three augmentation methods for two types of neural models
through six text classification tasks.
The corresponding code is implemented by Chainer~\cite{Tokui15} and available~\footnote{\url{https://github.com/pfnet-research/contextual_augmentation}}.

The benchmark datasets used are as follows: (1, 2) SST is a dataset for sentiment classification on movie reviews, which were annotated with five or two labels (SST5, SST2)~\cite{socher13}.
(3) Subjectivity dataset (Subj) was annotated with whether a sentence was subjective or objective~\cite{pang04}.
(4) MPQA is an opinion polarity detection dataset of short phrases rather than sentences~\cite{wiebe05}.
(5) RT is another movie review sentiment dataset~\cite{pang05}.
(6) TREC is a dataset for classification of the six question types (e.g., person, location)~\cite{li02}.
For a dataset without development data, we use 10\% of its training set for the validation set as well as Kim~\shortcite{kim14}.

We tested classifiers using the LSTM-RNN or CNN, and both have exhibited good performances.
We used typical architectures of classifiers based on the LSTM or CNN with dropout~\cite{hinton12}
using hyperparameters found in preliminary experiments.~\footnote{
An RNN-based classifier has a single layer LSTM and word embeddings, whose output is fed into an output affine layer with the softmax function.
A CNN-based classifier has convolutional filters of size \{3, 4, 5\} and word embeddings~\cite{kim14}. The concatenated output of all the filters are applied with a max-pooling over time and fed into a two-layer feed-forward network with ReLU, followed by the softmax function.
For both the architectures, training was performed by Adam~\cite{Kingma15} and
finished by early stopping with validation at each epoch.

The hyperparameters of the models and training were selected by a grid-search using baseline models without data augmentation in each task's validation set individually. We used the best settings from the combinations by changing the learning rate, unit or filter size, embedding dimension, and dropout ratio.
}
The reported accuracies of the models were averaged over eight models trained from different seeds.

The tested augmentation methods are: (1) synonym-based augmentation, and (2, 3) contextual augmentation with or without a label-conditional architecture.
The hyperparameters of the augmentation (temperature $\tau$ and probability of word replacement) were also selected by a grid-search using validation set, while retaining the hyperparameters of the models.
For contextual augmentation,
we first pretrained a bi-directional LSTM LM without the label-conditional architecture, on WikiText-103 corpus~\cite{Merity16} from a subset of English Wikipedia articles.
After the pretraining, the models are further trained on each labeled dataset with newly introduced label-conditional architectures.

\subsection{Results}

\setlength\tabcolsep{1.3pt}
\begin{table}
\small
\centering
\begin{tabular}{l|cccccc|c}
{\bf Models} &
{\bf STT5} &
{\bf STT2} &
{\bf Subj} &
{\bf MPQA} &
{\bf RT} &
{\bf TREC} &
{\bf Avg.}
\\ \hline \hline
CNN & 41.3 & 79.5 & 92.4 & 86.1 & 75.9 & 90.0 & 77.53 \\
w/ synonym & 40.7 & 80.0 & 92.4 & 86.3 & 76.0 & 89.6 & 77.50 \\
w/ context & 41.9 & 80.9 & 92.7 & 86.7 & 75.9 & 90.0 & 78.02 \\
\ \ \ \ \ + label & 42.1 & 80.8 & 93.0 & 86.7 & 76.1 & 90.5 & {\bf 78.20} \\
\hline
RNN & 40.2 & 80.3 & 92.4 & 86.0 & 76.7 & 89.0 & 77.43 \\
w/ synonym & 40.5 & 80.2 & 92.8 & 86.4 & 76.6 & 87.9 & 77.40 \\
w/ context & 40.9 & 79.3 & 92.8 & 86.4 & 77.0 & 89.3 & 77.62 \\
\ \ \ \ \ + label & 41.1 & 80.1 & 92.8 & 86.4 & 77.4 & 89.2 & {\bf 77.83} \\

\end{tabular}
\caption{\label{tab:results} Accuracies of the models for various benchmarks. The accuracies are averaged over eight models trained from different seeds.}
\end{table}

Table~\ref{tab:results} lists the accuracies of the models with or without augmentation.
The results show that our contextual augmentation improves the model performances for various datasets from different domains more significantly than the prior synonym-based augmentation does.
Furthermore, our label-conditional architecture boosted the performances on average and achieved the best accuracies.
Our methods are effective even for datasets with more than two types of labels, SST5 and TREC.

\begin{figure}[!t]
  \begin{center}
    \includegraphics[width=7.6cm,clip]{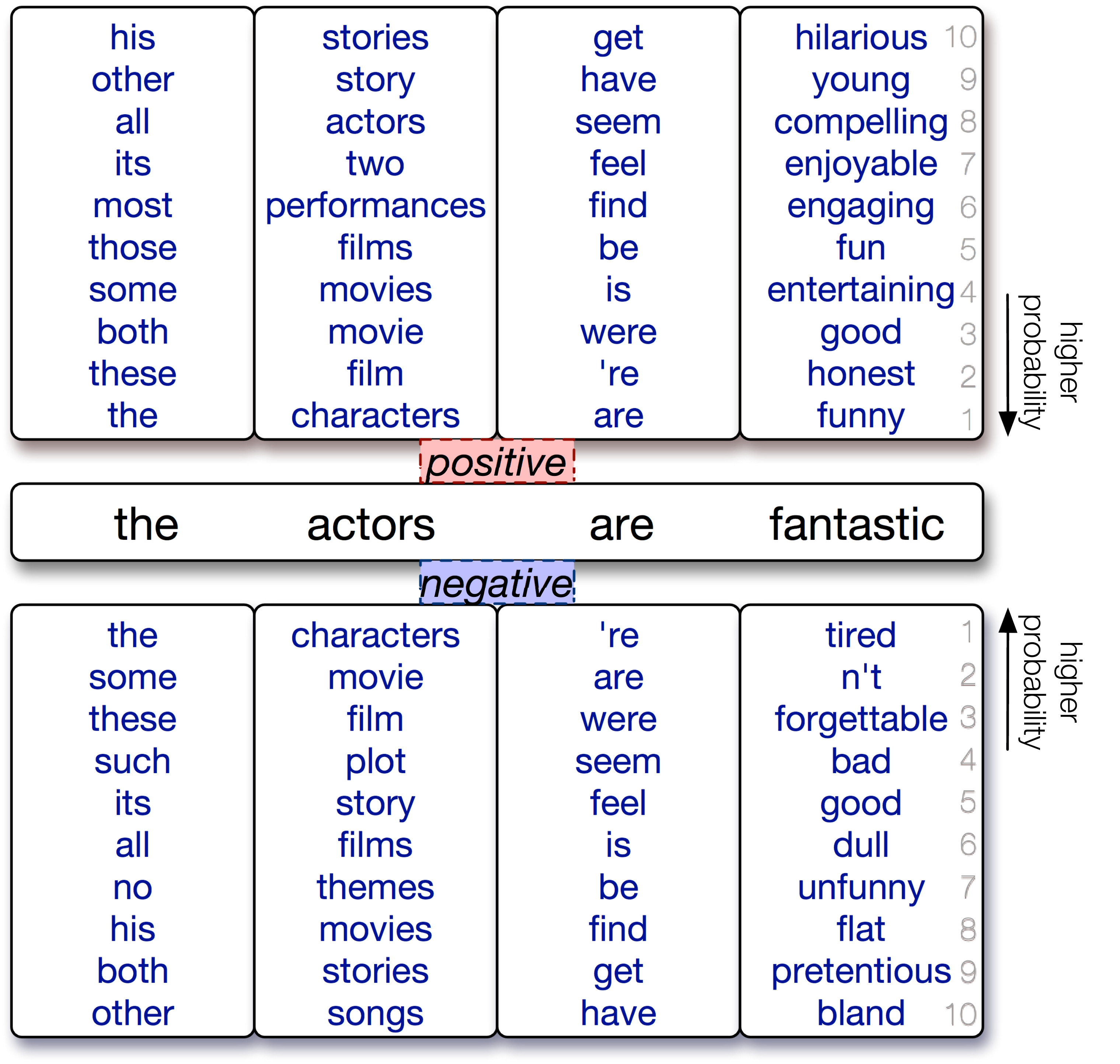}
    \caption{
    Words predicted with the ten highest probabilities by the conditional bi-directional LM applied to the sentence {\it ``the actors are fantastic''}. The squares above the sentence list the words predicted with a positive label. The squares below list the words predicted with a negative label.
}
    \label{fig:contextual_aug_examples}
  \end{center}
\end{figure}

For investigating our label-conditional bi-directional LM, we show in Figure~\ref{fig:contextual_aug_examples} the top-10 word predictions by the model for a sentence from the SST dataset.
Each word in the sentence is frequently replaced with various words that are not always synonyms.
We present two types of predictions depending on the label fed into the conditional LM.
With a positive label, the word ``fantastic'' is frequently replaced with {\it funny, honest, good,} and {\it entertaining}, which are also positive expressions.
In contrast, with a negative label, the word ``fantastic'' is frequently replaced with {\it tired, forgettable, bad}, and {\it dull}, which reflect a negative sentiment.
At another position, the word ``the'' can be replaced with ``no'' (with the seventh highest probability), so that the whole sentence becomes ``no actors are fantastic.'', which seems negative as a whole.
Aside from such inversions caused by labels,
the parts unrelated to the labels (e.g., ``actors'') are not very different in the positive or negative predictions.
These results also demonstrated that conditional architectures are effective.

\section{Related Work}

Some works tried text data augmentation
by using synonym lists~\cite{zhang15,wang15},
grammar induction~\cite{jia16},
task-specific heuristic rules~\cite{furstenau09,kafle17,silfverberg17},
or neural decoders of autoencoders~\cite{bergmanis17,xu17,hu17}
or encoder-decoder models~\cite{kim16,sennrich16back,xia17}.
The works most similar to our research are
Kolomiyets et al.~\shortcite{kolomiyets11} and Fadaee et al.~\shortcite{fadaee17}.
In a task of time expression recognition,
Kolomiyets et al.
replaced only the headwords
under a task-specific assumption that temporal trigger words usually occur as headwords.
They selected substitute words with top-K scores given by the Latent Words LM~\cite{deschacht09},
which is a LM based on fixed-length contexts.
Fadaee et al.~\shortcite{fadaee17}, focusing on the rare word problem in machine translation,
replaced words in a source sentence with only rare words,
which both of rightward and leftward LSTM LMs independently predict with top-K confidences.
A word in the translated sentence is also replaced
using a word alignment method and a rightward LM.
These two works share the idea of the usage of language models with our method.
We used a bi-directional LSTM LM which captures variable-length contexts with considering both the directions jointly.
More importantly, we proposed a label-conditional architecture and demonstrated its effect both qualitatively and quantitatively.
Our method is independent of any task-specific knowledge, and effective for classification tasks in various domains.

We use a label-conditional fill-in-the-blank context for data augmentation.
Neural models using the fill-in-the-blank context have been invested in other applications.
\citet{Kobayashi16,kobayashi17} proposed to extract and organize information about each entity in a discourse using the context.
\citet{fedus18} proposed GAN~\cite{goodfellow14} for text generation and demonstrated that the mode collapse and training instability can be relieved by in-filling-task training.
\citet{Melanmug16} and \citet{Peters18} reported that encoding the context with bidirectional LM was effective for a broad range of NLP tasks.

\section{Conclusion}

We proposed a novel data augmentation
using numerous words given by a bi-directional LM, and further introduced a label-conditional architecture into the LM. Experimentally, our method produced various words compatibly with the labels of original texts and improved neural classifiers more than the synonym-based augmentation.
Our method is independent of any task-specific knowledge or rules, and can be generally and easily used for classification tasks in various domains.

On the other hand, the improvement by our method is sometimes marginal.
Future work will explore comparison and combination with other generalization methods exploiting datasets deeply as well as our method.

\section*{Acknowledgments}

I would like to thank the members of Preferred Networks, Inc., especially Takeru Miyato and Yuta Tsuboi, for helpful comments. I would also like to thank anonymous reviewers for helpful comments.

\bibliography{naaclhlt2018}
\bibliographystyle{acl_natbib}

%

\end{document}